\documentclass[sigconf]{aamas} 

\usepackage{balance} %

\usepackage[ruled, linesnumbered, vlined]{algorithm2e} 
\usepackage{subcaption}
\usepackage{cleveref}

\setcopyright{none}
\acmConference[ALA '24]{Proc.\@ of the Adaptive and Learning Agents Workshop (ALA 2024)}
{May 6-7, 2024}{Online, \url{https://ala2024.github.io/}}{Avalos, Müller, Wang, Yates (eds.)}
\copyrightyear{2024}
\acmYear{2024}
\acmDOI{}
\acmPrice{}
\acmISBN{}
\title[CAESAR]{CAESAR: Enhancing Federated RL in Heterogeneous MDPs through Convergence-Aware Sampling with Screening}

\author{Hei Yi Mak}
\affiliation{
  \institution{ETH Zurich}
  \city{Zurich}
  \country{Switzerland}}
\email{heimak@ethz.ch}

\author{Flint Xiaofeng Fan}
\affiliation{
  \institution{National University of Singapore}
  \country{Singapore}}
\email{fxf@u.nus.edu}

\author{Luca A. Lanzendörfer}
\affiliation{
  \institution{ETH Zurich}
  \city{Zurich}
  \country{Switzerland}}
\email{lanzendoerfer@ethz.ch}

\author{Cheston Tan}
\affiliation{
  \institution{A*STAR}
  \country{Singapore}}
\email{cheston-tan@i2r.a-star.edu.sg}

\author{Wei Tsang Ooi}
\affiliation{
  \institution{National University of Singapore}
  \country{Singapore}}
\email{ooiwt@comp.nus.edu.sg}

\author{Roger Wattenhofer}
\affiliation{
  \institution{ETH Zurich}
  \city{Zurich}
  \country{Switzerland}}
\email{wattenhofer@ethz.ch}

\newcommand{\methodname}{\texttt{CAESAR}}
\newcommand{\schemeself}{\texttt{Self}}
\newcommand{\schemeall}{\texttt{All}}
\newcommand{\schemesame}{\texttt{Peers}}
\newcommand{\schemeadapt}{\texttt{Sampling}}
\newcommand{\schemescreen}{\texttt{Screen}}

\newcommand{\gridworld}{{\texttt{GridWorld}}}
\newcommand{\frozenlake}{{\texttt{FrozenLake-v1}}}

\begin{abstract}
In this study, we delve into Federated Reinforcement Learning (FedRL) in the context of value-based agents operating across diverse Markov Decision Processes (MDPs).  
Existing FedRL methods typically aggregate agents' learning by averaging the value functions across them to improve their performance.
However, this aggregation strategy is suboptimal in heterogeneous environments where agents converge to diverse optimal value functions.
To address this problem, we introduce the \textbf{C}onvergence-\textbf{A}war\textbf{E} \textbf{SA}mpling with sc\textbf{R}eening (\methodname{}) aggregation scheme
designed to enhance the learning of individual agents across varied MDPs.
\methodname{} is an aggregation strategy used by the server that combines convergence-aware sampling with a screening mechanism. 
By exploiting the fact 
that agents learning in identical MDPs are converging to the same optimal value function,
\methodname{} enables the selective assimilation of knowledge from more proficient counterparts, thereby significantly enhancing the overall learning efficiency.
 We empirically validate our hypothesis and 
 demonstrate the effectiveness
 of \methodname{} in enhancing the learning efficiency of agents, using
both a custom-built \gridworld{} environment and the classical \frozenlake{} task, each presenting varying levels of environmental heterogeneity.
\end{abstract}


\newcommand{\BibTeX}{\rm B\kern-.05em{\sc i\kern-.025em b}\kern-.08em\TeX}

\begin{document}

\pagestyle{fancy}
\fancyhead{}

\maketitle

\section{Introduction}

Federated Reinforcement Learning (FedRL)~\cite{frl-survey, frl-chapter} is a burgeoning field in Reinforcement Learning. Distinct for its collaborative learning approach, FedRL enables distributed agents to learn collectively while maintaining the privacy of their local data — the raw trajectories sampled from the local environments. FedRL leverages techniques in Federated Learning (FL), notably Federated Averaging~\cite{fed-avg}, to aggregate agent parameters to improve learning efficiency. 
{
While existing research on FedRL~\cite{frl-linear, frl-YQ, frl-fedhql, frl-pmlr-v202-woo23a,frl-fault-tolerant, frl-zhang2022resilient,DAI2024257-FRL-chapter} predominantly assumes \emph{homogeneous} environments, where all local environments correspond to the same Markov Decision Process (MDP)~\cite{rl-book} with identical dynamics and rewards, real-world applications often defy this assumption.
}
For instance, in the healthcare domain, FedRL holds promise for optimizing predictive models across various hospitals, each characterized by distinct patient demographics and disease patterns \cite{frl-clinical}. 
This diversity among patient populations and clinical manifestations leads to inherent \emph{heterogeneity} within the data environments shaped by the MDPs.

This challenge is underscored in the research by~\citet{frl-hetero-envs}.
While their work investigates FedRL in the context of heterogeneous environments, it primarily focuses on training a unified model to perform consistently across disparate local environments.
This approach, akin to implementing a standard healthcare protocol across hospitals serving diverse patient populations, may prove to be impractical.
Such a one-size-fits-all approach fails to accommodate the unique healthcare needs and specific disease prevalence of different communities, potentially resulting in suboptimal or even detrimental outcomes. This underscores the critical need for tailored approaches that respect and respond to  the unique characteristics of each environment.

In contrast, our research is centered on scenarios where each agent learns a localized policy for its designated MDP.
This is analogous to designing customized healthcare strategies for each hospital, taking into account the unique health demographics and local environmental influences of their patient population.
We explore the potential of these agents to collaboratively enhance the learning of localized policies, each specifically tailored to the corresponding environment. 
A pivotal assumption in our work is the unknown nature of both the number of distinct MDPs and the specific assignments of agents to these MDPs.
In response to these challenges, we propose a convergence-aware adaptive sampling strategy for value-based agents in FedRL settings characterized by heterogeneous environments. 
This strategy is based on the insight that value functions of agents optimizing for the same MDP are expected to converge towards a singular optimal value over time, 
thereby naturally reducing the variance in learning trajectories among these agents, or "\textit{peers.}"
Preliminary experiments suggest that while this strategy 
is effective in filtering out "\textit{non-peers}"—agents whose environmental contexts or MDPs diverge significantly from one another, leading to disparate optimal policies and value functions—it might inadvertently prioritize the inclusion of suboptimal peers.
These are agents within the same MDP whose strategies or learning progress are not as advanced, potentially anchoring the group to suboptimal points.
To address this, we introduce an additional screening process, aimed at incorporating only those agents that exhibit better performance. This dual approach of adaptive sampling and selective screening effectively mitigates the risk of suboptimal peer selection, enhancing the learning efficacy of agents in their respective MDPs.

In this paper, we address the challenges of training individual policies with environmental heterogeneity in FedRL. We begin by formulating the problem setup of FedRL in heterogeneous environments (Sec.~\ref{sec:problem-setup}) and proceed to examine various aggregation schemes (Sec.~\ref{sec:method}). We then introduce the \textbf{C}onvergence-\textbf{A}war\textbf{E} \textbf{SA}mpling with sc\textbf{R}eening (\methodname{}) aggregation scheme that tailors the average value functions for individuals to effectively improve their learning (Sec.~\ref{subsec:caesar}).
\methodname{} stands out for its dual-layered approach: 
firstly, utilizing a convergence-aware sampling mechanism for efficient peer identification in diverse MDPs (Sec.~\ref{subsec:sampling}); 
and
secondly, 
incorporating a selective screening process (Sec.~\ref{subsec:screen}) to refine agent interactions, prioritizing only those agents that demonstrate superior performance.
We empirically validate the effectiveness and robustness of \methodname{} to improve agents' learning using environments of \gridworld{} and \frozenlake{}, 
engineered for the purpose of illustrating environmental heterogeneity (Sec.~\ref{sec:eval}).
We have made our work publicly available and open-sourced,\footnote{\url{https://github.com/hughiemak/CAESAR}}
providing new perspectives and viable approaches for tackling the challenges of FedRL in heterogeneous settings.

\section{Preliminaries}\label{sec:prelim}

\textbf{Markov Decision Processes (MDPs).}
In the realm of reinforcement learning, sequential decision-making problems are commonly modeled using MDPs~\cite{rl-book}. An MDP is characterized by a $6$-tuple $(\mathcal{S}, \mathcal{A}, \mathcal{P},\mathcal{R},\gamma, \rho)$ where $\mathcal{S}$ denotes the state space, $\mathcal{A}$ represents the action space, $\mathcal{P}(s'|s,a)$ defines the transition probabilities between states, $\mathcal{R}:\mathcal{S}\times \mathcal{A}\rightarrow \mathbb{R}$ is the reward function, $\gamma$ is the discount factor, and $\rho$ is the initial state distribution.

\textbf{Q-learning.}
Q-learning~\cite{qlearn} stands as a cornerstone in classical reinforcement learning, operating as an off-policy temporal difference algorithm.
In Q-learning, an agent learns an action-value function $Q: \mathcal{S} \times \mathcal{A}\rightarrow \mathcal{R}$ using a table. The entry $Q(s,a)$, also known as the Q-values, estimates the expected return of taking action $a\in \mathcal{A}$ in state $s\in \mathcal{S}$. The value of $Q(s,a)$ is updated by applying the Bellman equation:
\begin{eqnarray}
Q(s,a) \leftarrow (1-\alpha) Q(s,a)+\alpha(r+\gamma\max_{a'}Q(s',a'))
\end{eqnarray}
where $s$ is the current state, $a$ is the current action to be executed, $r$ is the immediate reward, $s'$ is the next state, and $\alpha$ is the learning rate.
Then a decision policy $\pi_Q$ can be obtained via {exploiting} the updated $Q$-values:
\begin{equation}
\label{eq:action:selection:greedy}
    \pi_Q(s) \leftarrow a_t = \arg \max_a Q(s_t, a) .
\end{equation}
The \emph{optimal action} at state $s_t$ is defined as $a^*_t = \arg \max_a Q^*(s_t, a)$
where $Q^*(s_t, a)$ is the \emph{optimal Q-function} which gives the expected return for starting in state $s_t$, taking action $a$, and following the policy thereafter.

\textbf{Federated Reinforcement Learning (FedRL).} Initially introduced by \citet{frl-YQ}, Federated Reinforcement Learning (FedRL) has gained increasing prominence, evidenced by its extensive application in various real-world scenarios \citep{FedRL-2,FedRL-1,yu2020deep-FedRL-3,wang2020federated-FedRL-5,FedRL-building,liang2019federated-FedRL-Car-YQ,frl-zhang2022resilient, frl-fedhql} and its substantial theoretical development \citep{frl-fault-tolerant, frl-hetero-envs, frl-linear, frl-pmlr-v202-woo23a,frl-jordan2024decentralized}. Notably, \citet{frl-fault-tolerant} conducted pioneering work on the robust convergence of federated policy gradients, demonstrating sublinear speedup.
\citet{frl-linear,frl-pmlr-v202-woo23a} further advanced the field by showcasing linear speedup in federated Q-learning under Markovian Sampling.
\citet{fedrl-shen2023towards} established a linear speedup for federated Actor-Critic algorithms under i.i.d. sampling. 
A common assumption in these related works is the homogeneity of MDPs across all agents participating in FedRL.
This perspective was expanded by
\citet{frl-hetero-envs}, who explored FedRL in the context of environmental heterogeneity. Their research primarily aimed at developing a global shared policy model within an imaginary MDP framework. 

\section{Federated Reinforcement Learning with Heterogeneous Environments}\label{sec:problem-setup}
\textbf{MDP Configuration.}
In the FedRL setting under consideration, we have $N$ agents, and a collection of MDPs, where the quantity of distinct MDPs ($K$) is less than or equal to $N$. Each MDP, denoted as $M_k$, shares a common state space $\mathcal{S}$ and action space $\mathcal{A}$, but is uniquely defined by its transition dynamics $\mathcal{P}_k(s'|s,a)$ and reward function $\mathcal{R}_k:\mathcal{S}\times \mathcal{A}\rightarrow \mathbb{R}$.
Thus, the MDP $M_k$ is represented by the 6-tuple $(\mathcal{S}, \mathcal{A}, \mathcal{P}_k,\mathcal{R}_k,\gamma, \rho_k)$, where $\gamma$ is the discount factor and $\rho_k$ is the initial state distribution specific to MDP $M_k$.
An assignment function $f: [N]\rightarrow [K]$ determines the allocation of each agent to these MDPs.

\begin{figure}[b]
  \centering
  \includegraphics[width=1.\linewidth]{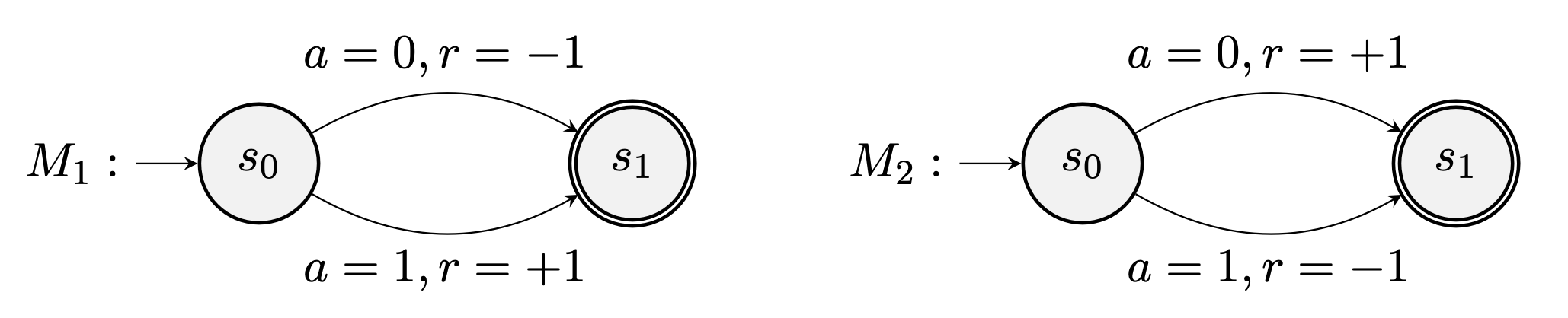}
  \caption{Two heterogeneous MDPs. MDP $M_1$ rewards $-1$ for action $0$ and $+1$ for action $1$, while MDP $M_2$ rewards $+1$ for action $0$ and $-1$ for action $1$. The optimal value functions are $Q_1(s_0,0)=-1, Q_1(s_0,1)=1$ for $M_1$, and $Q_2(s_0,0)=1, Q_2(s_0,1)=-1$ for $M_2$, respectively. Averaging these value functions results in $\bar{Q}(s_0,0) = \bar{Q}(s_0,1) = 0$, showing a misrepresentation of optimal values for both MDPs.}
  \label{fig:all-problem}
\end{figure}

\textbf{Heterogeneity in MDPs.}
The core of heterogeneity in this setting stems from the differences in transition dynamics and reward functions among the MDPs. An example of this heterogeneity is depicted in Fig.~\ref{fig:all-problem}, where two MDPs share the same state and action spaces but have distinct reward functions. This diversity in dynamics and rewards exemplifies the complexity and variability agents encounter in heterogeneous environments.

\textbf{Operational Assumptions.}
A pivotal assumption in our approach is the unknown nature of both $K$ (the number of MDPs) and the assignment function $f$.
This uncertainty adds a layer of complexity to the learning process, as agents must navigate and adapt to their assigned MDPs without prior knowledge of the overall system configuration.

\textbf{Agent Learning and Objectives.}
Each agent in our system, denoted as $i$, is a value-based learner, employing techniques such as Q-learning for policy optimization.
Every Agent $i$ interacts solely with a local instantiation of its assigned MDP, $M_{f(i)}$, from which it gathers and analyzes sample trajectories to inform its learning. 
The primary goal for each agent is to optimize its action-value function $Q_i$, aiming to achieve optimal expected performance within its unique local environment.
This focus on individual optimization within a shared learning framework underscores the challenge of balancing local adaptation with collaborative learning in FedRL.

\textbf{Federated Updates in FedRL.}
Following existing FedRL work~\cite{frl-fault-tolerant, frl-linear, frl-YQ, frl-fedhql}, a central server is available to coordinate the federated learning. 
We consider a FedRL training process where a federated update takes place every $H$ local updates. 
At each local update step, each agent performs a standard learning step using the value-based RL algorithm. 
During the federated update, for each agent $i$, the server will select a subset of agents $S\subseteq [N]$ and aggregate their value functions into a new value function $\bar{Q}$. 
For tabular Q-learning, a straightforward aggregation method is averaging the value functions (Q-tables) across selected agents:
\begin{eqnarray}\label{eq:fedQ-avg}
    \bar{Q}(s,a)=\sum_{j\in S} Q_j(s,a), \quad \forall s\in \mathcal{S}, a\in \mathcal{A}.
\end{eqnarray}
After aggregation, the server sends $\bar{Q}$ to agent $i$ and updates $Q_i$ towards $\bar{Q}$:
\begin{equation}\label{eq:fed-update}
    Q_i(s,a)\leftarrow \beta Q_i(s,a) + (1-\beta) \bar{Q}(s,a),\quad \forall s\in \mathcal{S}, a\in \mathcal{A}
\end{equation}
where $\beta \in [0,1]$ is a blending parameter controlling the extent of update from the federated value function. 

A key challenge in the federated update process is determining the optimal subset $S$ for each agent without prior knowledge of $f$, the agent's specific environment, or direct access to its local trajectories. 
The selection of $S$ is pivotal in ensuring that the aggregated value function $\bar{Q}$ is conducive to agent $i$'s learning in its MDP.
In Sec.~\ref{sec:method}, we will explore various aggregation schemes for selecting $S$.

\section{Aggregation Schemes}\label{sec:method}
In this section, we explore various schemes for selecting the subset of agents, $S$, for each agent $i$, culminating in the introduction of our novel \methodname{} scheme.

\subsection{\schemeself{}}
The \schemeself{} scheme serves as a baseline, where agents learn independently, without federated updates.
Using this scheme,
Eq.~\eqref{eq:fedQ-avg} can be viewed as:
\begin{eqnarray*}
    \bar{Q} = Q_i, \ \  \forall i \in N
\end{eqnarray*}
implying no external influence during the federated update phase. 
Consequently, the selected subset $S$ only includes the agent itself:
\begin{eqnarray*}
    S=\{i\}.
\end{eqnarray*}
It is a fundamental expectation that, for agents to be incentivized to engage in the federative process, any employed selection scheme must ensure that the aggregated knowledge surpasses the performance achievable by the \schemeself{}. This is essential to justify the collaborative effort in the federative learning context.

\subsection{\schemeall{}}\label{subsec:all}
The scheme \schemeall{} is another baseline corresponding to the canonical FedRL averaging scheme where all agents are included for aggregation to compute Eq.~\eqref{eq:fedQ-avg}:
\begin{eqnarray*}
    \bar{Q}(s,a)=\sum_{j=1}^N Q_j(s,a).
\end{eqnarray*}
In this scheme, the selected subset always includes all agents:
\begin{eqnarray*}
    S=\{1,2,\dots, N\}.
\end{eqnarray*}
In a FedRL setting characterized by heterogeneous local environments, the \schemeall{} scheme may impede the learning process and potentially obstruct convergence.
This issue arises because each agent's value function, denoted as $Q_j$, is being optimized for different MDPs.
In essence, they are converging towards disparate optimal value functions.
Consequently, value functions optimized for one MDP might adversely affect the aggregated value function $\bar{Q}$, resulting in misleading guidance for agent $i$.
To illustrate, consider the scenario with two simple MDPs, $M_1$ and $M_2$, as shown in Fig.~\ref{fig:all-problem}. Suppose agents 1 and 2 are learning in these MDPs respectively and have both reached their optimal value functions: $Q_1(s_0,0)=-1, Q_1(s_0,1)=1$ for MDP $M_1$, and $Q_2(s_0,0)=1, Q_2(s_0,1)=-1$ for MDP $M_2$. However, the averaged Q-values, $\bar{Q}(s_0,0)$ and $\bar{Q}(s_0,1)$, both result in 0. These average values are suboptimal for both MDPs. Updating $Q_1$ and $Q_2$ based on $\bar{Q}$ would therefore misguide the agents and steer them away from their currently optimal values, highlighting the challenge of aggregation in heterogeneous environments.

\subsection{\schemesame{}}\label{subsec:peers}
The \schemesame{} scheme is an unrealistic scheme in our setting that serves as a hypothetical benchmark. This scheme operates under the assumption of having prior knowledge of MDP assignments, denoted as $f$, and including only those agents assigned to the same MDP as agent $i$. These agents are referred to as the `peers' of agent $i$. 
The selected subset of agents is therefore
\begin{eqnarray*}
    S=\{j\in [N]: f(i)=f(j) \}.
\end{eqnarray*}
Such a presumption renders it impractical in scenarios where this information is not available, i.e., the server lacks insight into the peers of agent $i$. 
Despite this, the scheme serves as a valuable benchmark, illustrating the potential advantages of precise, environment-specific aggregation, such that:
\begin{eqnarray*}
    \bar{Q}(s,a)=\sum_{j\in S} Q_j(s,a), \ \  S=\{j\in [N]: f(i)=f(j) \}.
\end{eqnarray*}

Contrasting with the \schemeall{} scheme (Sec.~\ref{subsec:all}), this conceptual approach offers greater efficiency by exclusively incorporating value functions that are optimized for the same MDP. This selective aggregation ensures that value functions from disparate MDPs, which could potentially mislead the learning process, are not included.
Furthermore, this scheme provides a distinct advantage over the \schemeself{} scheme, where agents learn in isolation. By leveraging the collective knowledge of agents assigned to the same MDP, it enables a more targeted and effective aggregation of value functions, enhancing the overall learning effectiveness.

\subsection{\schemeadapt{}}\label{subsec:sampling}
Inspired by the advantageous attributes of the hypothetical \schemesame{} scheme, we explore the feasibility of devising a similar selection scheme.
Our goal is to accurately identify the peers of Agent $i$ without relying on the prior assumption of peer knowledge inherent to the \schemesame{} approach.
This task is especially challenging in our scenario due to the lack of prior knowledge about each agent's assigned MDP and the absence of direct access to local trajectories at the server. 

To circumvent this, we propose utilizing the convergence of agent value functions as a heuristic for peer detection.
The main idea is that if the value functions $Q_i$ and $Q_j$ are both being optimized for the same MDP, they should converge towards a unique optimal value function $Q^*$ over time. 
As a result, the values $Q_i(s,a)$ and $Q_j(s,a)$ for all state-action pairs $(s,a)$ will progressively become more similar. 
We empirically validate this convergence behavior in a gridworld setting, as detailed in Fig.~\ref{fig:gridworld-converge} (Sec.~\ref{subsec:verification}).

Given this intuition, the convergence of value functions emerges as a practical heuristic for estimating whether two agents are learning in the same MDP. 
This insight leads to our convergence-aware sampling scheme, \schemeadapt{}, wherein the subset $S$ is sampled based on probabilities $p_{i1}, \dots, p_{iN}$.
Each probability $p_{ij}$ quantifies the likelihood of including agent $j$ in $S$ and is dynamically adjusted based on the observed convergence between $Q_i$ and $Q_j$.
At the onset of training, the server initializes an $N \times N$ matrix $p$, where the entry $p_{ij}$ is set as:
\begin{align}
p_{ij}=\begin{cases}p_0&i\neq j,\\ 1&i=j.\end{cases}, \quad \forall i, j\in[N]
\end{align}
where $p_0$ functions as an initial assumption or `prior' about the task, reflecting the preliminary likelihood of agents being peers before any learning occurs. By default, it can be assigned a value of $0$ to encourage self-learning at the start of training when the convergence information is insufficient.
Prior to each federated update, 
the server updates the entries $p_{ij} $ of the probability matrix $p$.
This update is contingent upon evaluating the evolving similarity between the value functions $Q_i$ and $Q_j$.
Specifically, the server assesses how the similarity of these value functions has changed relative to their states observed $H$ steps ago.
This dissimilarity between two value functions $Q$ and $Q'$ is defined as the mean absolute difference across all state-action pairs:
\begin{align*}
    d(Q, Q')=\frac{1}{|\mathcal{S}|\times |\mathcal{A}|}\sum_{s,a} |Q(s,a)-Q'(s,a)|.
\end{align*}
Let $Q_k^{(t)}$ be agent $k$'s current value function and $Q_k^{(t-H)}$ be agent $k$'s value function $H$ steps ago. For each pair of agents $\{i,j\}$, we update
\begin{equation*}
    p_{ij}\leftarrow \begin{cases}
                    \min(p_{ij}+\delta, 1) & \text{if }d(Q_i^{(t-H)}, Q_j^{(t-H)}) - d(Q_i^{(t)}, Q_j^{(t)}) > \xi,\\
                    \max(p_{ij}-\delta, 0) &\text{otherwise.}
                \end{cases}
\end{equation*}
where $\delta > 0, \xi \geq 0$. 
This update rule is designed such that if the value functions $Q_i$ and $Q_j$ demonstrate a sufficient decrease in dissimilarity over a specific time window $H$, the server will increase the probability value  $p_{ij}$. This increment in $p_{ij}$ effectively raises the likelihood of agent $j$ being selected for agent $i$'s subset for aggregation.
The time window $H$ acts as a temporal frame of reference, enabling the server to assess changes in similarity over a defined period.
Conversely, if $Q_i$ and $Q_j$ do not exhibit the required degree of convergence over the time window, $p_{ij}$ is reduced. 
Persistent convergence trends lead to a gradual increment in $p_{ij}$, favoring the selection of agents with converging value functions.
Consequently, the \schemeadapt{} scheme dynamically adapts its selection criteria over time, increasingly favoring the inclusion of agents with value functions that demonstrate a tendency to converge.
Parameters $\delta$ and $\xi$ control the sensitivity of $p_{ij}$ adjustments and the required degree of convergence, respectively.

\SetArgSty{textrm}
\begin{algorithm}[]
\caption{\methodname{}}\label{alg:fedrl}
    \DontPrintSemicolon
    \SetKwProg{Fn}{}{:}{}
    \SetKwFunction{execute}{ServerExecutes}
    \SetKwFunction{localupdate}{LocalUpdate}
    \SetKwFunction{fedupdate}{FederatedUpdate}
    \SetKwFunction{updatep}{UpdatePMatrix}
    \SetKwFunction{evallocalperformance}{EvalLocalPerformance}
    \Fn{\execute{$H, \sigma, p_0, \delta, \xi, \beta$}}{
        initialize value functions $Q_i$ for each agent $i\in [N]$\;
        initialize $Q_i^{old} \leftarrow Q_i$ for each agent $i\in [N]$\;
        initialize matrix $p$: $p_{ij}=\begin{cases}p_0&i\neq j,\\ 1&i=j.\end{cases}, \forall i, j\in[N]$\;
        \For{each step $t=1,\dots, T$}{
            \localupdate{$i$} for each agent $i\in [N]$\;
            \If{$t$ mod $H=0$}{
                $g_k \leftarrow $ \evallocalperformance{$k$}, $\forall k\in [N]$\;
                \updatep{$p, \delta, \xi, \{Q_k^{old}\}_k, \{Q_k\}_k$}\;
                \fedupdate{$i, \beta, p, \{Q_k\}_k, \{g_k\}_k$} for each agent $i\in [N]$\;
                $Q_i^{old}\leftarrow Q_i$ for each agent $i\in [N]$\;
            }
        }
    }\;
    \Fn{\updatep{$p, \delta, \xi, \{Q_k^{old}\}_k, \{Q_k\}_k$}}{
        \For{agent $i=1$ to $N$}{
            \For{agent $j=i+1$ to $N$}{
                $p_{ij}\leftarrow \begin{cases}
                    \min(p_{ij}+\delta, 1) & \text{if } d(Q_i^{old}, Q_j^{old}) - d(Q_i, Q_j) > \xi,\\
                    \max(p_{ij}-\delta, 0) &\text{otherwise.}
                \end{cases}$\;
            }
        }
    }\;
    \Fn{\fedupdate{$i, \beta, p, \{Q_k\}_k, \{g_k\}_k$}}{
        initialize $S'=\{\}$\;
        \For{$j\in [N]$}{
            add $j$ to $S'$ with probability $p_{ij}$
        }
        $S=\{j: j\in S' \text{ and } g_j > g_i\}$\;
        Construct $\bar{Q}$: $\bar{Q}(s,a)=\sum_{j\in S} Q_j(s,a), \quad \forall s\in \mathcal{S}, a\in \mathcal{A}$\;
        Update $Q_i$: $Q_i(s,a)\leftarrow \beta Q_i(s,a) + (1-\beta) \bar{Q}(s,a),\quad \forall s\in \mathcal{S}, a\in \mathcal{A}$
        
    }
\end{algorithm}
\subsection{\methodname{}}\label{subsec:caesar}
As will be discussed in Sec.~\ref{sec:eval}, the \schemeadapt{} scheme excels at filtering out non-peers from the set $S$, but it also has a potential downside: the server might inadvertently incorporate only peers that are underperforming, confining slow-progressing agents in suboptimal points in the value function space. To mitigate this issue, we introduce an additional screening process to refine agent interactions, prioritizing only those agents that demonstrate superior performance, culminating in the \methodname{} aggregation scheme.

In the \methodname{} scheme, we initially select a subset of agents based on the probabilities $p_{i1},\dots, p_{iN}$, following the same process as in \schemeadapt{}.
The primary objective of this sampling step, akin to that in \schemeadapt{}, is to identify probable peers by assessing the convergence trends of their value functions. Subsequently, we introduce a screening layer, which focuses on the comparative performance of these selected agents. The rationale behind this additional step is to circumvent the pitfall of updating the value function $Q_i$ towards the average of lower-performing peers, which could hinder the convergence of $Q_i$ to its optimal state.

To implement this, the local performance of each agent $k$, $g_k\approx \mathbb{E}_{M_k, \pi_{Q_k}}\left [\sum_{t=0}^\infty \gamma^t \mathcal{R}_k(s_t,a_t) \right]$, is measured prior to the federated update.
During the update, for a given agent $i$, the server initially samples a preliminary subset of agents, $S^{\prime}$, in line with the probabilities $p_{i1},\dots, p_{iN}$.
It then further refines $S^{\prime}$ by retaining only those agents whose performance, $g_j$, exceeds that of agent $i$ $(g_j > g_i)$. The final subset for aggregation is thus defined as: 
\begin{eqnarray*}
    S=\{j: j\in S' \text{ and } g_j > g_i\}
\end{eqnarray*}

This resulting subset $S$ is then utilized to assemble the aggregated value function $\bar{Q}$ for updating $Q_i$ of each agent. 
This completes the outlines for the \methodname{} scheme. For a detailed procedural breakdown, refer to the pseudocode presented in Algorithm~\ref{alg:fedrl}.

\subsection{\schemescreen{}}\label{subsec:screen}
As a complementary approach, the \schemescreen{} scheme focuses solely on the screening process based on local performance, without considering convergence trends:
\begin{eqnarray*}
    S=\{j: g_j > g_i\}
\end{eqnarray*}
\schemescreen{}
selects agents that are performing better than the target agent, but may include those from different MDPs. This scheme tests the efficacy of performance-based selection in isolation.

Each scheme presents a unique approach to aggregating value functions within a FedRL framework. Our goal, as detailed in Sec.~\ref{sec:eval}, is to assess the effectiveness of these schemes in enhancing individual agent performances, particularly in heterogeneous environments.
This analytical endeavor aims to uncover the most effective strategies for knowledge aggregation in practical FedRL settings, thereby providing valuable insights into optimizing agent performance in diverse and complex scenarios.

\section{Empirical Evaluation}\label{sec:eval}

\subsection{Experimental Settings}
In this study, we conduct a comparative analysis of the six aggregation schemes discussed in Sec.~\ref{sec:method}. 
For this comparison, we employ Q-learning agents within two distinct environments: a custom-built environment \gridworld{} and the well-known \frozenlake{} task from the OpenAI Gym toolkit~\cite{open-ai}.

\begin{figure}[h]
  \centering
  \includegraphics[width=1.\linewidth]{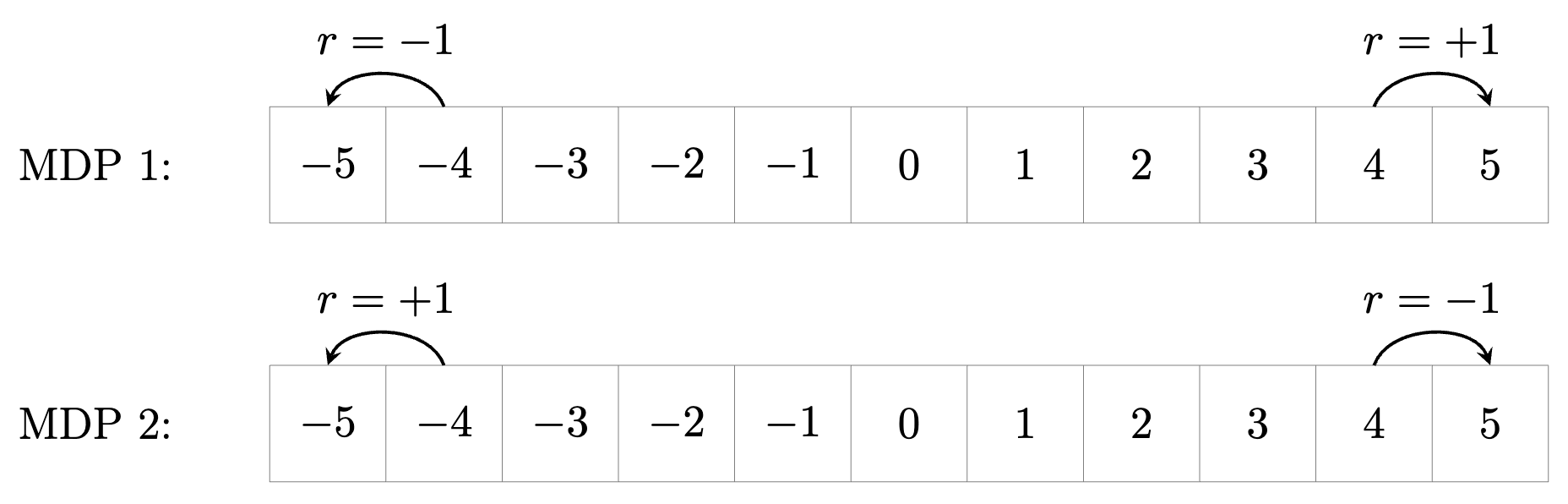}
  \caption{Two \gridworld{} MDPs. Their initial states are $0$. In MDP 1 (top), transiting from state $4$ to $5$ generates a reward of $+1$ and transiting from state $-4$ to $-5$ yields a reward of $-1$. In MDP 2 (bottom), the signs of the rewards are flipped.}\label{fig:gridworlds}
  \Description{The two \gridworld{} MDPs. The initial state is $0$ in both MDPs. In MDP 1 (top), transiting from state $4$ to $5$ generates a reward of $-1$ and transiting from state $-4$ to $-5$ yields a reward of $-1$. In MDP 2 (bottom), the signs of the rewards are flipped.}
\end{figure}

\begin{figure}[t]
    \centering
    \begin{subfigure}[h]{0.15\textwidth}

        \centering
        \includegraphics[width=2cm]{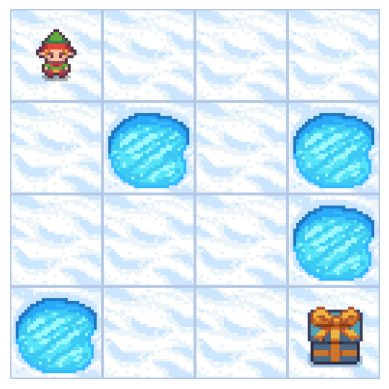}
        \caption{Map 0}
        \label{fig:frozenlake-map0}
    \end{subfigure}
    \begin{subfigure}[h]{0.15\textwidth}
        \centering
        \includegraphics[width=2cm]{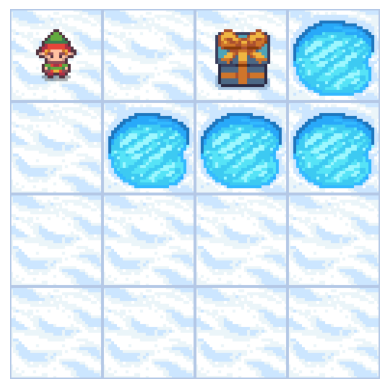}
        \caption{Map 1}
        \label{fig:frozenlake-map1}
    \end{subfigure}
    \begin{subfigure}[h]{0.15\textwidth}
        \centering
        \includegraphics[width=2cm]{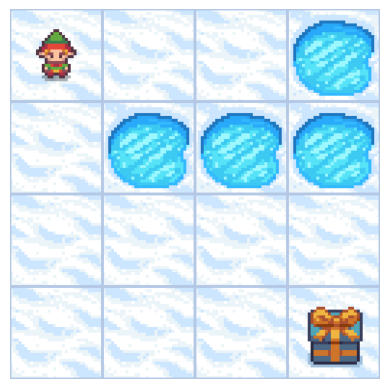}
        \caption{Map 2}
        \label{fig:frozenlake-map2}
    \end{subfigure}
    \caption{\frozenlake{} environments generated by three different maps. 
  The agent’s task is to navigate to the goal (the gift box) without falling into the holes.}
    \label{fig:frozenlake}
    \Description{\frozenlake{} environments generated by three different maps. 
  The agent’s task is to navigate to the goal (the gift box) without falling into the holes.}
\end{figure}

The \gridworld{} is designed as a 1-dimensional discrete environment, characterized by a state space $\mathcal{S}=\{-5,-4,\dots,4,5\}$ and a binary action space $\mathcal{A}=\{0,1\}$. The initial state for each episode is set at $0$, with terminal states being $5$ and $-5$. The agent's actions impact the state transitions: action $0$ moves the state from $x$ to $x-1$, while action $1$ advances the state from $x$ to $x+1$.
Two distinct versions of this environment are considered, corresponding to two different MDPs.
Fig.~\ref{fig:gridworlds} provides a visual representation of these \gridworld{} environments.
In the first MDP (MDP 1), a transition from state $4$ to $5$ yields a reward of $+1$, and a transition from $-4$ to $-5$ results in a reward of $-1$. All other state transitions provide a neutral reward of $+0$. The second MDP (MDP 2) inverts the reward structure of MDP 1, such that $r_2(s,a)=-r_1(s,a)$ for all $s\in \mathcal{S}$ and $ a\in \mathcal{A}$). 
The \frozenlake{} environment presents a 2-dimensional discrete challenge that effectively encapsulates the complexities of environmental heterogeneity.
In this environment, agents are tasked with navigating to a designated goal while avoiding hazardous holes.
The environment is characterized by a four-directional action space, and episodes end with a reward of $+1$ upon reaching the goal, or $+0$ if the agent falls into a hole or exhausts the allowed steps.
The heterogeneity of the \frozenlake{} environment is induced by the distinct map configurations, as shown in Fig.~\ref{fig:frozenlake}.
Each map represents a unique instantiation of a local MDP within the environment, characterized by its own specific arrangement of holes and paths, 
necessitating different strategic approaches for successful navigation.
This diversity in maps 
provides a practical scenario to assess how FedRL algorithms perform across dynamically varied MDPs.

\subsection{Hypothesis Verification Using \gridworld{}}\label{subsec:verification}
For \gridworld{}, we partition $N=20$ agents into $K=2$ groups, each comprising $10$ agents.
These groups are then assigned to two different MDPs, $M_1$ and $M_2$, as depicted in Fig.~\ref{fig:gridworlds}.
Each agent is trained for $T=10000$ steps with an exploration rate $\epsilon=0.1$, and receives a federated update every $H=100$ steps.

\begin{figure}[h]
  \centering
  \includegraphics[width=0.9\linewidth]{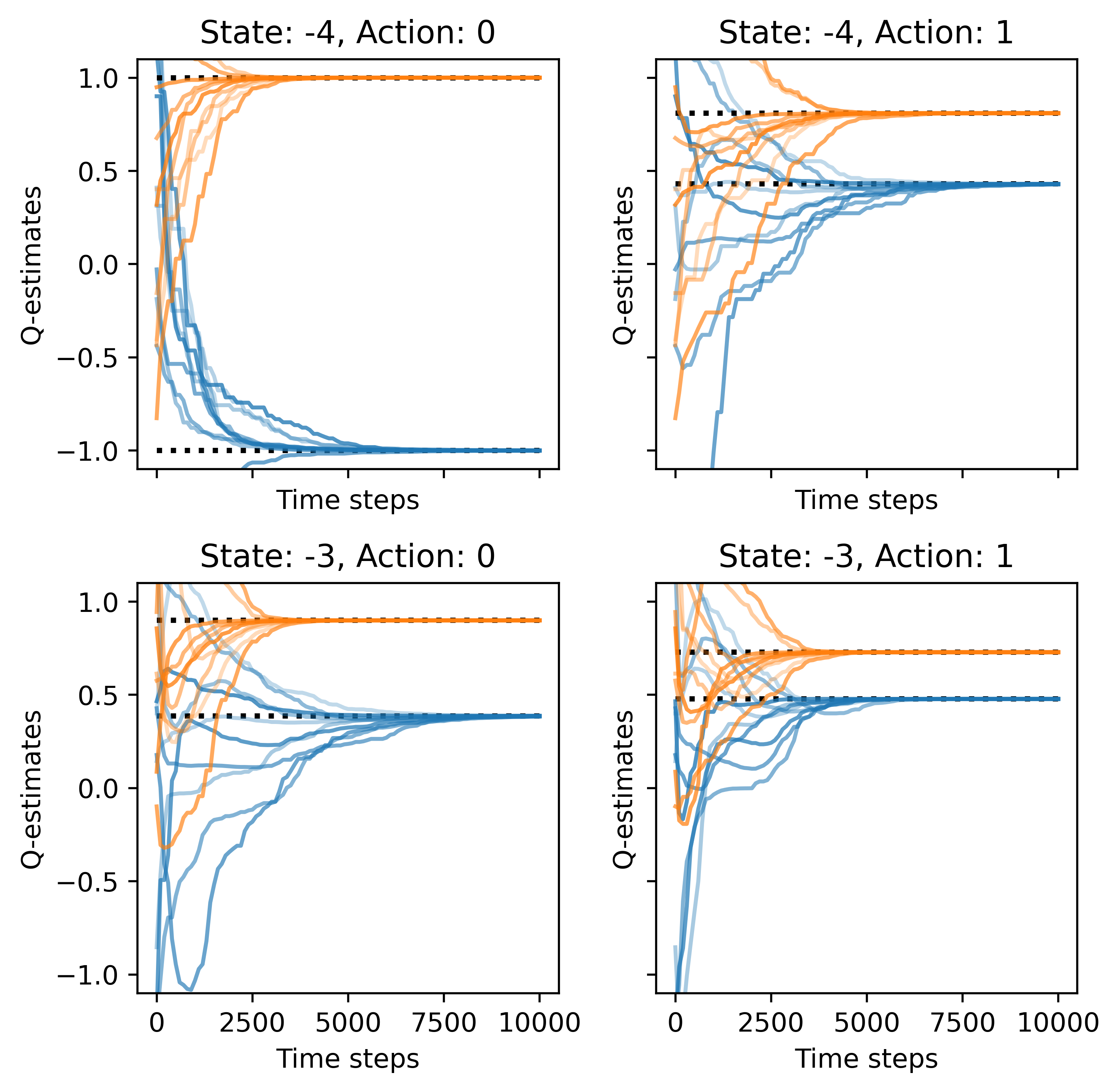}
  \caption{Convergence of Q-values among peers in \gridworld{} under \schemeself{}. Q-values of $M_1$ agents (blue) and $M_2$ agents (orange) converge to their respective optimal values (black dotted lines) for state-actions $(s=-4,a=\cdot)$ and $(s=-3,a=\cdot)$ in \gridworld{}. $\epsilon$ is set to $0.9$ to speed up convergence.}
\label{fig:gridworld-converge}
  \Description{Convergence of Q-values among peers in \gridworld{} under \schemeself{}. Q-values of $M_1$ agents (blue) and $M_2$ agents (orange) converge to their respective optimal values (black dotted lines) for state-actions $(s=-4,a=\cdot)$ and $(s=-3,a=\cdot)$ in \gridworld{}. $\epsilon$ is set to $0.9$ to speed up convergence.}
\end{figure}

\textbf{Convergence Among Peers.}
In the relatively simple \gridworld{} environment, 
we capture the agents' Q-tables every $H$ steps. 
Fig.~\ref{fig:gridworld-converge} shows how Q-values $Q_i(s,a)$ for various state-action $(s, a)$ pairs evolve for all agents $i \in [N]$ under  \schemeself{} (independent learning).
The optimal values of these state-action pairs are different for $M_1$ and $M_2$.
Notably, Q-values among peers converge towards the optimal values for their respective MDPs over time, supporting the use of Q-value convergence as a heuristic for detecting probable peers, as elaborated in Sec.~\ref{subsec:sampling}.

\begin{figure}[t]
    \centering
    \includegraphics[width=0.7\linewidth]{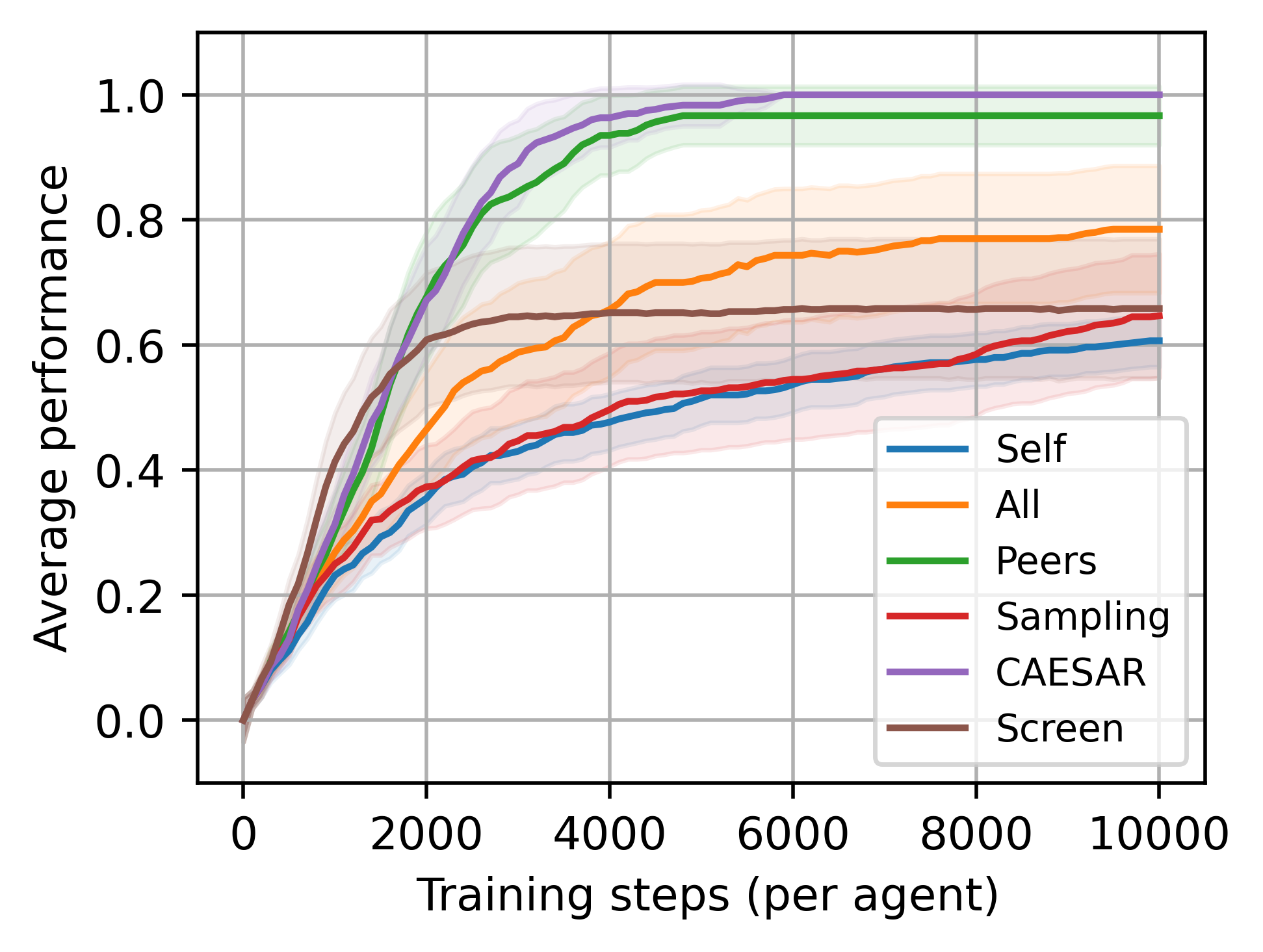}
    \caption{Average performance of the $N=20$ agents in \gridworld{} under different averaging schemes with exploration rate $\epsilon=0.1$. 
    The plot averages independent runs over 30 random seeds where the shadows represent the $95\%$ confidence intervals.}
    \label{fig:gridworld-performance}
    \Description{Average performance of the $N=20$ agents in \gridworld{} under different averaging schemes with exploration rate $\epsilon=0.1$.}
\end{figure}

\textbf{Comparative Performance Analysis.}
Fig.~\ref{fig:gridworld-performance} illustrates the average performance (over 30 random seeds) of all agents under different aggregation schemes in \gridworld{}.
We can observe that \schemeall{} is outperformed by \schemesame{}, affirming our hypothesis that including all agents in $S$ to compute $\bar{Q}$ according to Eq.~\eqref{eq:fedQ-avg}
is less effective in heterogeneous environments. 
The slower learning progress observed under \schemescreen{} is attributed to its selection based solely on local performance, often including high-performing agents from different MDPs.
Significantly, \methodname{} shows comparable results to the hypothetical approach \schemesame{}, which operates under the assumption of perfect knowledge about agent-MDP assignments. 
Remarkably, \methodname{} surpasses both \schemeadapt{} and \schemescreen{}, highlighting the synergistic effect of their combination on learning enhancement.

\begin{figure*}[ht!]
  \includegraphics[width=\textwidth]{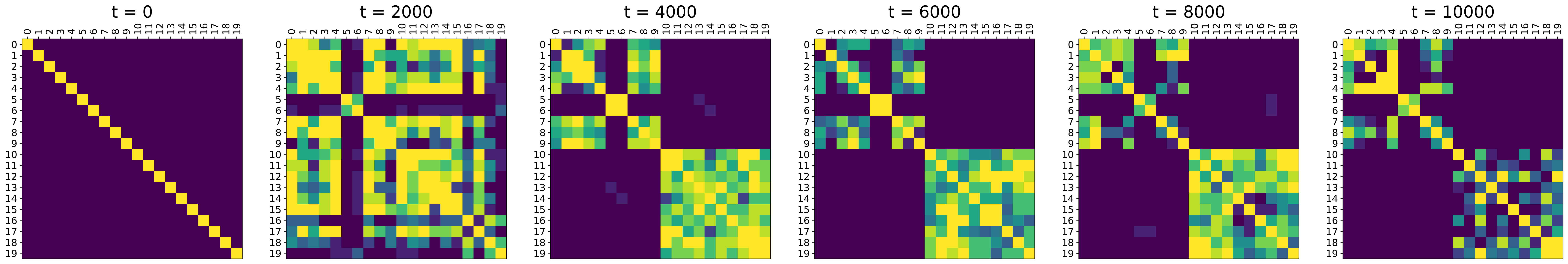}
  \caption{The matrix $p$ as a heatmap (yellow and purple indicate $1$ and $0$ respectively) at $5$ different time points under \schemeadapt{}. The numbers on the axes correspond the agents, where agents $0$ to $9$ are assigned to $M_1$ and agents $10$ to $19$ are assigned to $M_2$. The color of the cell $(i,j)$ indicates the probability of selecting agent $j$ for agent $i$.}
  \label{fig:gridworld-p-matrix-sampling-seed3}
  \Description{The matrix $p$ as a heatmap (yellow and purple indicate $1$ and $0$ respectively) at $5$ different time points under \schemeadapt{}. The numbers on the axes correspond to the agents, where agents $0$ to $9$ are assigned to $M_1$ and agents $10$ to $19$ are assigned to $M_2$. The color of the cell $(i,j)$ indicates the probability of selecting agent $j$ for agent $i$.}
\end{figure*}

\begin{figure}[h]
    \centering
    \includegraphics[width=0.9\linewidth]{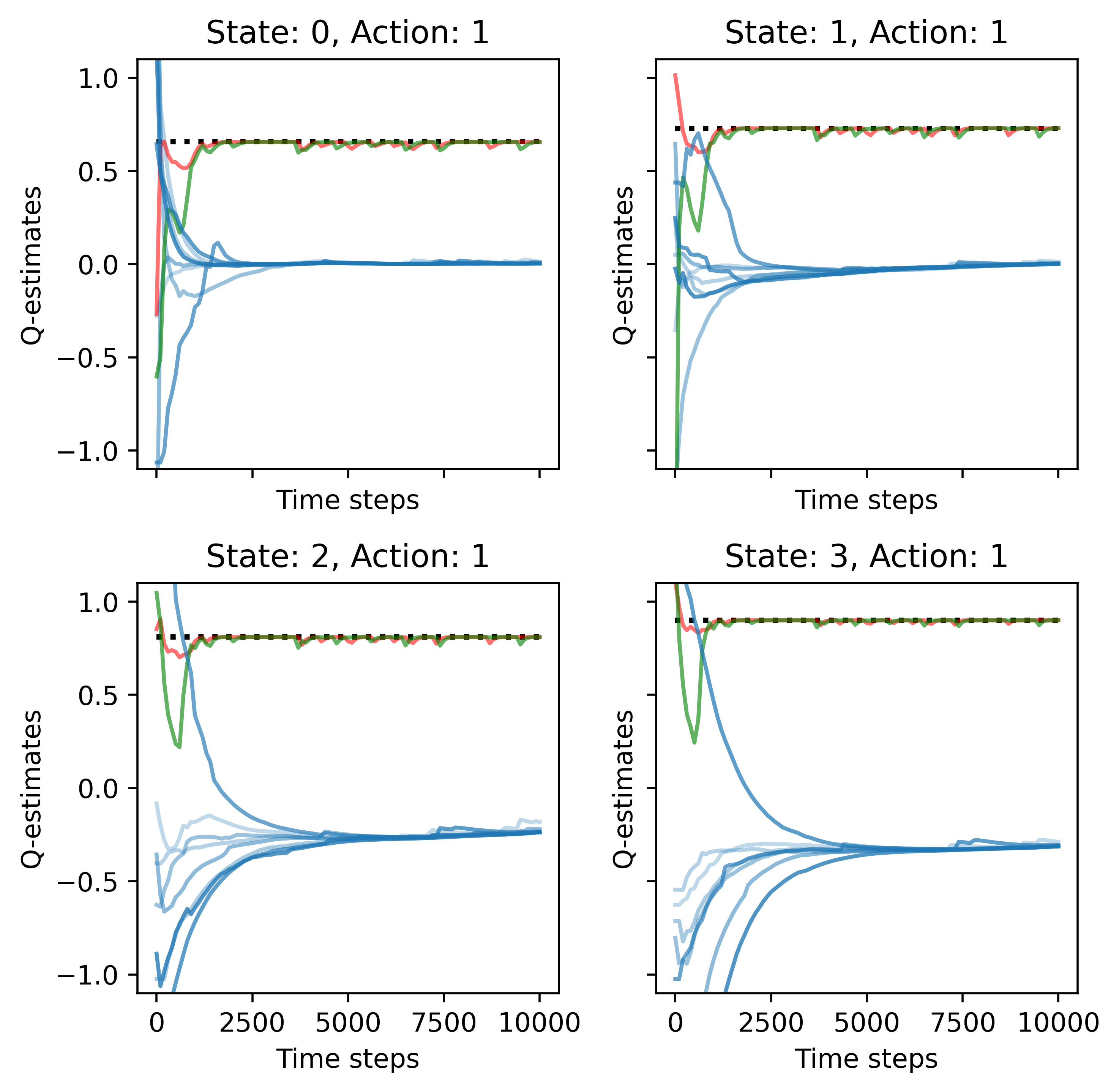}
    \caption{Q-values of the $M_1$ agents under \schemeadapt{}. 
    Two $M_1$ agents, Agent 5 (red curve) and Agent 6 (green curve), exhibit fast learning progress and converge to the true optimal values (black dotted lines) but the remaining $M_1$ agents (blue curves), Agents 0, 1, 2, 3, 4, 7, 8, 9, converge to non-optimal values.}
    \label{fig:gridworld-q-values-sampling-seed3}
    \Description{Q-values of the $M_1$ agents under \schemeadapt{}. 
    Two $M_1$ agents, Agent 5 (red curve) and Agent 6 (green curve), exhibit fast learning progress and converge to the true optimal values (black dotted lines) but the remaining $M_1$ agents (blue curves), Agents 0, 1, 2, 3, 4, 7, 8, 9, converge to non-optimal values.}
\end{figure}

\begin{figure}[h]
    \centering
    \includegraphics[width=0.9\linewidth]{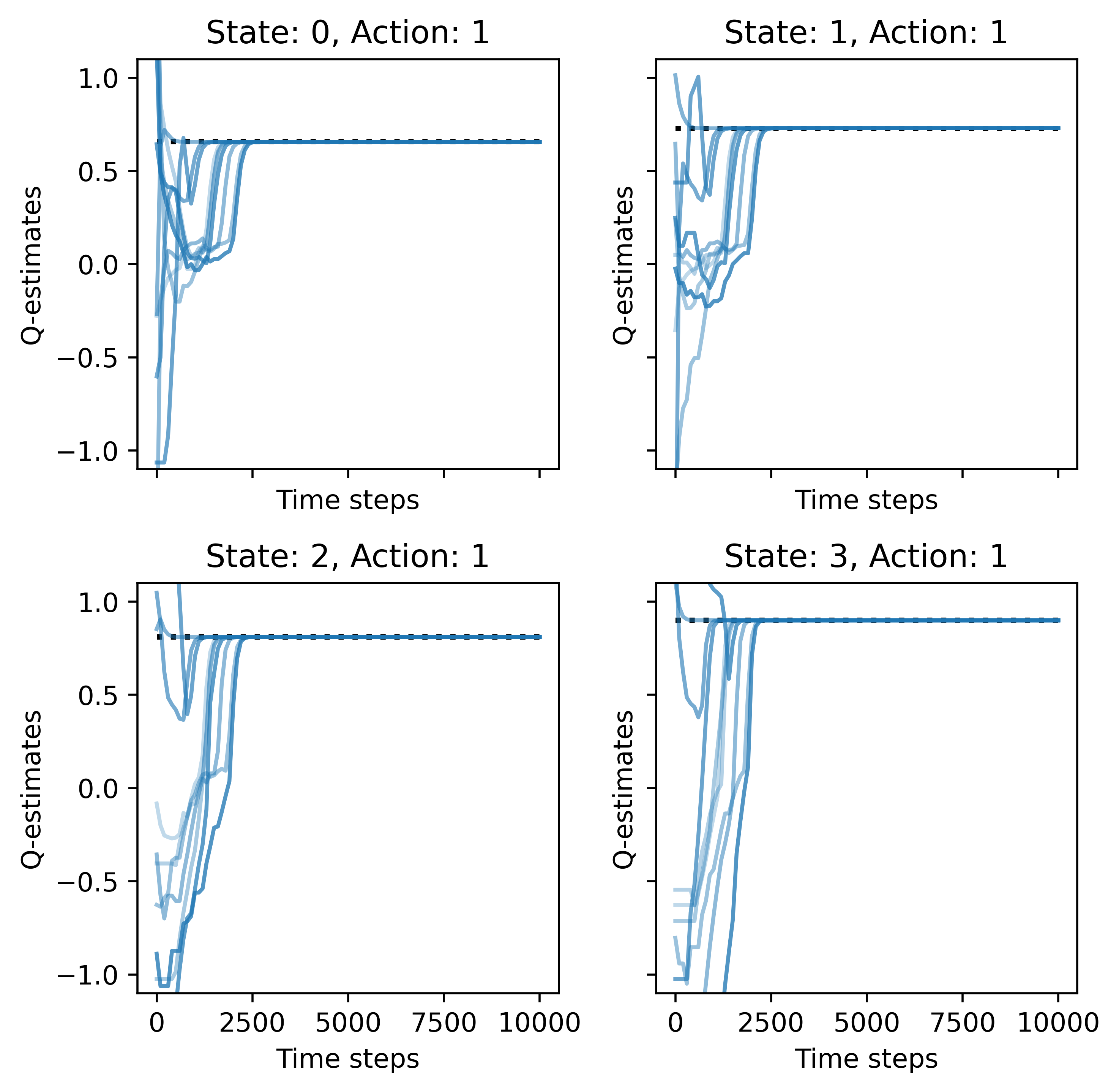}
    \caption{Under \methodname{}, Q-values of all $M_1$ agents converge quickly to the true optimal values (black dotted lines).}
    \label{fig:gridworld-q-values-caesar-seed3}
    \Description{Under \methodname{}, Q-values of all $M_1$ agents converge quickly to the true optimal values (black dotted lines).}
\end{figure}

\textbf{Analysis of Q-Value Evolution.}
To understand the critical role of the screening process in the \methodname{} scheme, we track the progression of Q-values throughout the training period.
Fig.~\ref{fig:gridworld-q-values-sampling-seed3} and Fig.~\ref{fig:gridworld-q-values-caesar-seed3} present the evolution of Q-values for agents assigned to $M_1$ under the \schemeadapt{} and \methodname{} schemes, respectively.
These plots are generated from training sessions with the same random seed and initial agent configurations.
Under \schemeadapt{}, we notice that only two agents are able to approximate the optimal Q-values (indicated by black dotted lines), while the remaining agents stagnate at suboptimal points. In contrast, when employing \methodname{}, a uniform and rapid convergence to optimal values is observed for all agents.

\begin{figure*}[h]
    \centering
    \begin{subfigure}[b]{0.33 \textwidth}
        \centering
        \includegraphics[width=\textwidth]{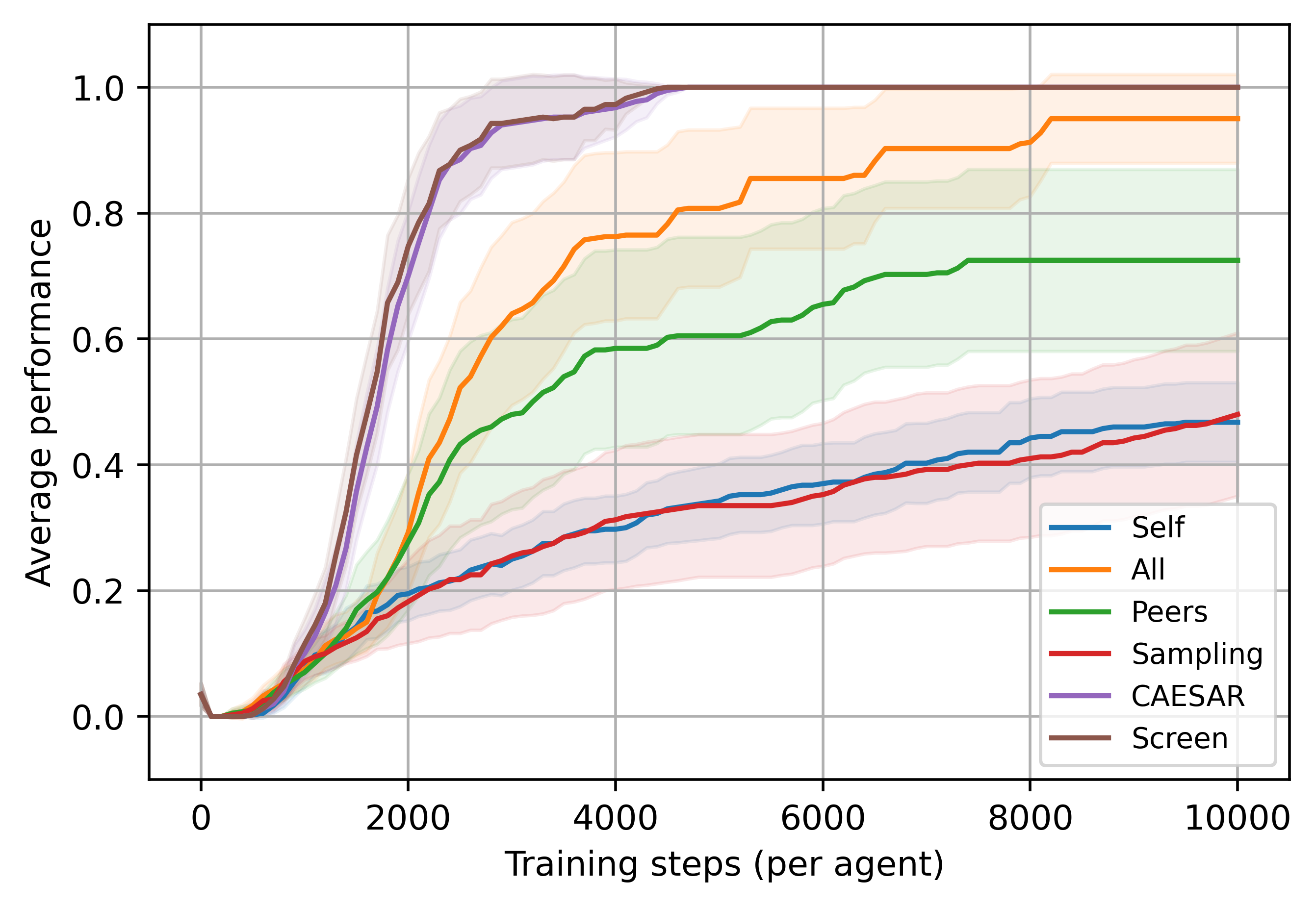}
        \caption{Homogeneous environments}
        \label{fig:frozenlake-homo}
    \end{subfigure}
    \begin{subfigure}[b]{0.33\textwidth}
        \centering
        \includegraphics[width=\textwidth]{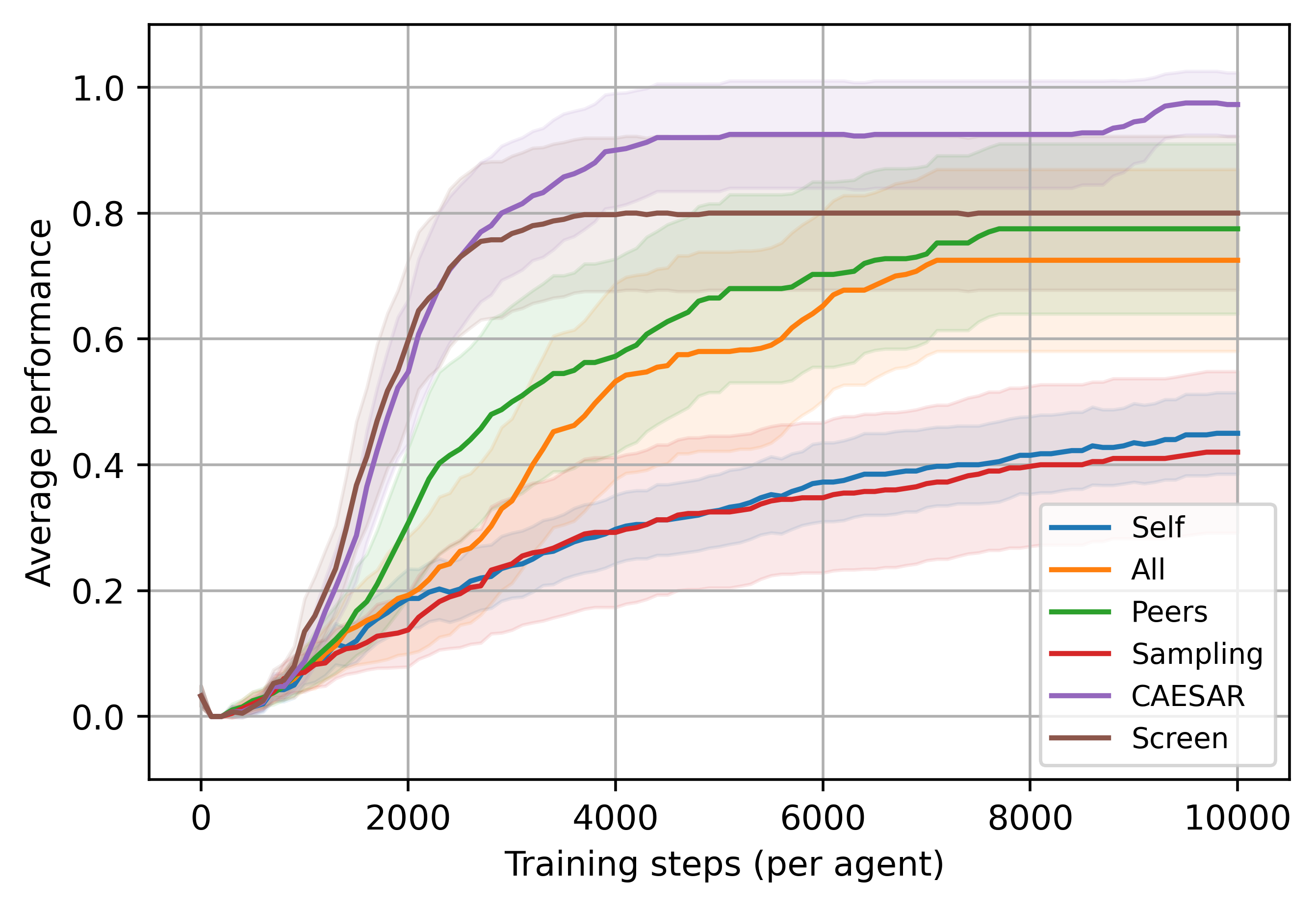}
        \caption{Random heterogeneous environments}
        \label{fig:frozenlake-hetero}
    \end{subfigure}
    \begin{subfigure}[b]{0.33\textwidth}
        \centering
        \includegraphics[width=\textwidth]{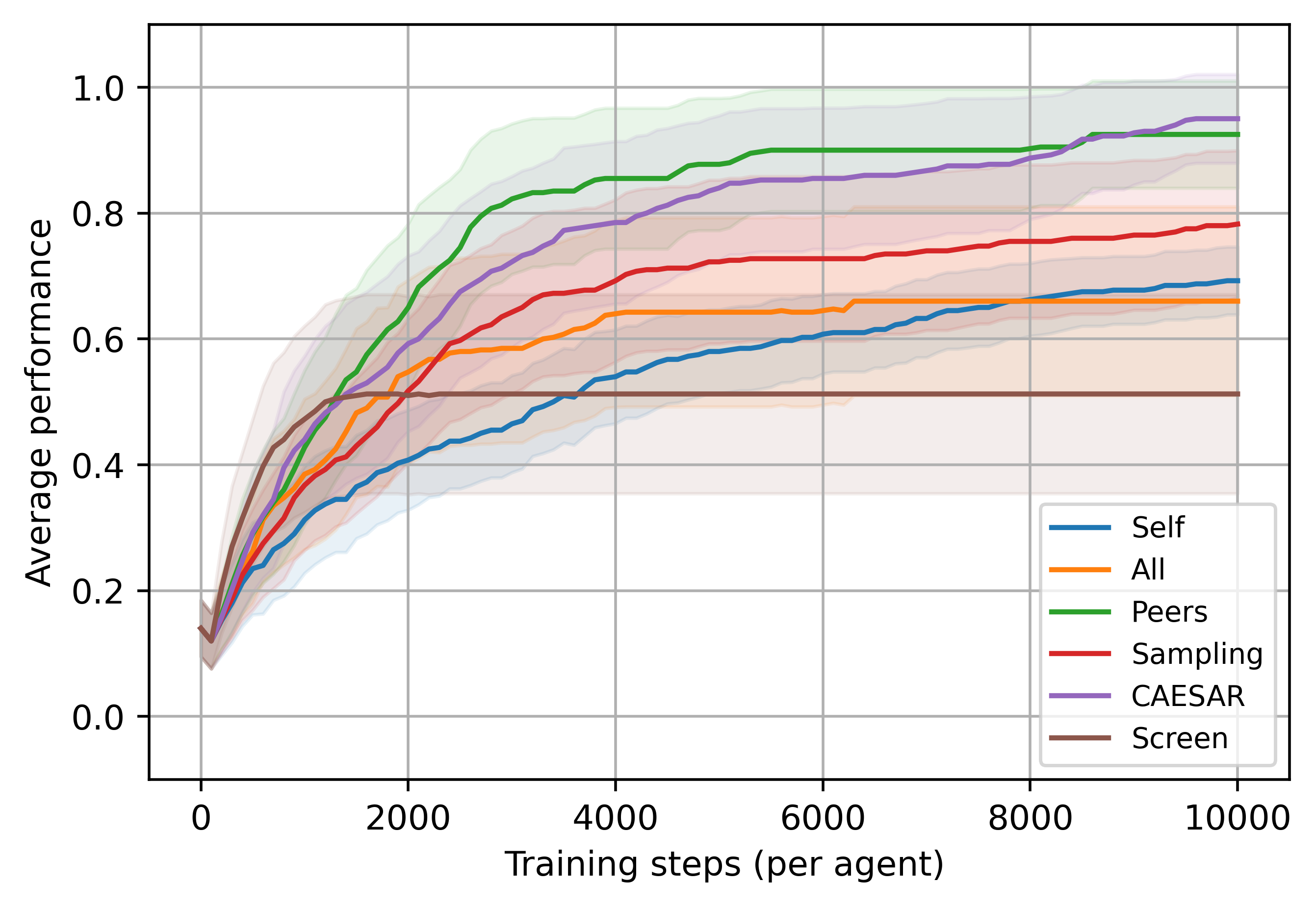}
        \caption{Strongly heterogeneous environments}
        \label{fig:frozenlake-strongly-hetero}
    \end{subfigure}
    \caption{Average performance of the $N=20$ agents in \frozenlake{} under different averaging schemes with exploration $\epsilon=0.1$ in all three settings. The plots average independent runs over 30 random seeds where the shadows represent the $95\%$ confidence intervals.}
    \label{fig:frozenlake-performance}
    \Description{Average performance of the $N=20$ agents in \frozenlake{} under different averaging schemes with exploration $\epsilon=0.1$ in all three settings. The plots average independent runs over 30 random seeds where the shadows represent the $95\%$ confidence intervals.}
\end{figure*}

To gain a deeper understanding of the dynamics at play within the \schemeadapt{} scheme, we analyze the changes in the $p$-matrix (Sec.~\ref{subsec:sampling}) over various training stages, as illustrated in Fig.~\ref{fig:gridworld-p-matrix-sampling-seed3}.
This analysis reveals that \schemeadapt{} is highly effective in filtering out non-peers, 
consistently selecting them with near-zero probability from timestep $t=4000$ onwards.
However, an intriguing behavior is observed: \schemeadapt{} tends to overlook agents who are advancing quickly in their learning curve, opting instead for peers with slower progress rates.
Specifically, at $t = 4000$ (as shown in the third plot of Fig.~\ref{fig:gridworld-p-matrix-sampling-seed3}), \schemeadapt{} assigns negligible probabilities
to aggregate values of the fast learners, Agents $5$ and $6$,
into the learning process of the slower-progressing peers, namely Agents $0, 1, 2, 3, 4, 7, 8, 9$, as shown in Fig.~\ref{fig:gridworld-q-values-sampling-seed3}.
Despite the evident progress of the fast learners, \schemeadapt{} scheme leads to a tendency for slower peers to primarily learn from each other, gravitating towards a consensus that strays from the optimal value.
Such a strategy, while fostering a form of convergence, risks cementing the learning of slower-progressing agents around suboptimal values.

In contrast, \methodname{} circumvents this issue through its dual-layered approach: it employs \schemeadapt{} to effectively identify and exclude non-peers, and \schemescreen{}, the screening process that prioritizes the inclusion of fast-progressing agents based on their local performance metrics, as evident in Fig.~\ref{fig:gridworld-q-values-caesar-seed3}.
This strategic selection tends to aggregate knowledge from faster-progressing peers, whose value functions $Q_i$ are more optimal for the same MDP.
Hence, \methodname{} not only avoids the pitfalls of \schemeadapt{} but also facilitates a more effective knowledge transfer, significantly enhancing the learning efficiency across agents.

\subsection{Effectiveness evaluation using \frozenlake{}}
In our study using \frozenlake{}, we maintain the same experimental settings as in \gridworld{}, with $N=20$ agents divided into two groups $K=2$, each group comprising 10 agents assigned to two distinct MDPs $M_1$ and $M_2$, respectively.
We assess the performance of the aggregation schemes under the following scenarios, each offering a different level of environmental heterogeneity:

\begin{enumerate}
    \item \textbf{Homogeneous environments:} $M_1$ and $M_2$ are identical, both generated using the same random map with 4 holes. An example of such a map is shown in Fig.~\ref{fig:frozenlake-map0} (a).
    \item \textbf{Randomly heterogeneous environments} $M_1$ and $M_2$ are distinct, created using two random maps with differing positions of the 4 holes.
    \item \textbf{Strongly heterogeneous environments} Maps 1 and 2, as depicted in Fig.~\ref{fig:frozenlake-map0} (b) and (c) respectively, exhibit a significant disparity in difficulty levels, with Map 1 being the easier and Map 2 the more challenging. The two maps are designed to have substantial differences in their optimal Q-functions.
\end{enumerate}

Fig.~\ref{fig:frozenlake-performance} shows the average performance of all agents across the different aggregation schemes in these scenarios. 
The results reveal that \methodname{} consistently demonstrates robust performance in all three scenarios, contrasting with other schemes that struggle in at least one scenario.

In Scenario 1 (Fig.~\ref{fig:frozenlake-homo}), with identical MDPs $M_1=M_2$, \schemeall{} demonstrates superior learning outcomes compared to \schemesame{}. 
This aligns with expectations, as all agents are engaged in the same task, making the inclusion of the entire agent pool in $S$ more effective for leveraging collective insights. In this context, \schemeall{} benefits from a broader knowledge base than \schemesame{}, which limits its focus to peers, hence reducing the number of participating agents.

Conversely, in Scenario 2 (Fig.~\ref{fig:frozenlake-hetero}), where 
  $M_1$ and $M_2$ differ, \schemesame{} slightly outperforms \schemeall{}. This indicates that peer-based learning is more advantageous when agents are dealing with different MDPs, as it enables more targeted knowledge sharing.

The contrast becomes more pronounced in Scenario 3 (Fig.~\ref{fig:frozenlake-strongly-hetero}), where \schemesame{} significantly surpasses \schemeall{}, with the latter even falling behind \schemeself{} (independent learning). 
This scenario underscores the importance of excluding non-peers from $S$ in heterogeneous environments. 
The discrepancy arises from the varying difficulty levels of $M_1$ (Map 1) and $M_2$(Map 2).
Agents in the simpler $M_1$ quickly master the task, leading to high-value estimates of $Q(s_0, a=\rightarrow)$ which is not optimal for $M_2$ where the action $a=\rightarrow$ often leads to holes.
Therefore, including $M_1$ agents' value functions in $S$ can detrimentally affect the learning progress of $M_2$ agents, as their optimal Q-value functions diverge significantly.

\schemescreen{} displays a notably inconsistent performance pattern across different levels of heterogeneity, excelling in Scenarios 1 and 2 but faltering in Scenario 3, where its results are even inferior to those of the independent learning approach, \schemeself{}. 
This phenomenon stems from that the \schemescreen{} scheme is effectively the \schemeall{} scheme with an additional screening process. In homogeneous environments (Scenario 1), where $M_1=M_2$, this screening process effectively boosts performance by prioritizing agents with superior performance.  
However, in the more complex Scenario 3, \schemescreen{} tends to erroneously include high-performing agents from the simpler $M_1$, whose optimal values are counterproductive in $M_2$.
As a result, \schemescreen{} inadvertently hinders the learning process for $M_2$ agents by propagating suboptimal Q-values.
This issue is clearly demonstrated in Fig.~\ref{fig:frozenlake-strongly-hetero}, where \schemescreen{} achieves an average performance of only 0.5, suggesting that half of the agents are unable to effectively address their assigned tasks.

\methodname{} demonstrates remarkable robustness across all three scenarios. Notably, in Scenario 1 (Fig.~\ref{fig:frozenlake-homo}), where $M_1$ and $M_2$ are identical and thus all agents are peers, the inclusion of the sampling process within \methodname{} does not impede learning gains. This is evidenced by its performance being on par with \schemescreen{}, suggesting that the additional process does not detract from learning efficiency in homogeneous environments.
In scenarios where $M_1$ and $M_2$ differ, particularly in the more complex Scenario 3, \methodname{} continues to show strong performance, in stark contrast to the diminishing results of \schemescreen{} and \schemeall{}. This resilience is primarily attributed to the sampling process integral to \methodname{}, which effectively filters out non-peers, thereby ensuring that agents are exposed to relevant and beneficial strategies for their specific environments. 
It is important to note that while \schemesame{} excels in scenario 3, its implementation is not practical in real applications where the agent-MDP assignments are not known, as discussed in Sec.~\ref{subsec:peers}.
\methodname{} stands out in practical settings where the degree of heterogeneity among environments might be unknown or unpredictable.
Its consistent performance across diverse scenarios underscores its suitability as a versatile and reliable aggregation strategy for FedRL in a practical setting.

\section{Conclusion}
In this study, we have tackled the intricate challenge of training distinct policies for agents across diverse environments within the realm of Federated Reinforcement Learning. 
Our investigation entailed a thorough analysis of six different aggregation strategies within the FedRL paradigm.

The experiments conducted in both customized \gridworld{} and \frozenlake{} demonstrated the efficacy of Q-value convergence as a heuristic for peer detection in FedRL. 
Notably, the proposed \methodname{} scheme stood out for its adaptability and resilience across a spectrum of environmental heterogeneity, consistently surpassing other evaluated baselines.
This adaptability makes \methodname{} particularly advantageous for real-world FedRL applications, where the unique characteristics of each environment are accommodated.

While our exploration focused on a tabular setting, future research directions include extending our methodologies to more complex and dynamic environments, especially those featuring a continuous control space. 
Furthermore, this work is primarily centered around agents that employ Q-value-based strategies. Acknowledging this as a limitation, another valuable direction for future research would be the incorporation of policy-based methods.

\bibliographystyle{ACM-Reference-Format} 
\bibliography{AAMAS_2024_CAESAR}

\end{document}